\def\year{2019}\relax
\documentclass[letterpaper]{article} %DO NOT CHANGE THIS
\usepackage{aaai19}  %Required
\usepackage{times}  %Required
\usepackage{helvet}  %Required
\usepackage{courier}  %Required
\usepackage{url}  %Required
\usepackage{graphicx}  %Required
\usepackage{natbib}

\begin{document}
\title{TransferTransfo: A Transfer Learning Approach for Neural Network Based Conversational Agents}
\author{Thomas Wolf, Victor Sanh, Julien Chaumond \& Clement Delangue \\
HuggingFace  Inc.\\
81 Prospect St.\\
Brooklyn, New York 11201, USA\\
{\tt \{thomas,victor,julien,clement\}@huggingface.co} \\}

\maketitle
\begin{abstract}
  We introduce a new approach to generative data-driven dialogue systems (e.g. chatbots) called \textit{TransferTransfo} which is a combination of a \textit{Transfer} learning based training scheme and a high-capacity \textit{Transfo}-rmer model. Fine-tuning is performed by using a multi-task objective which combines several unsupervised prediction tasks. The resulting fine-tuned model shows strong improvements over the current state-of-the-art end-to-end conversational models like memory augmented seq2seq and information-retrieval models. On the privately held {\sc persona-chat} dataset of the Conversational Intelligence Challenge 2, this approach obtains a new state-of-the-art, respectively pushing the perplexity, Hits@1 and F1 metrics to $16.28$ ($45\%$ absolute improvement), $80.7$ ($46\%$ absolute improvement) and $19.5$ ($20\%$ absolute improvement).
\end{abstract}

\section{Introduction}

Non-goal-oriented dialogue systems (chatbots) are an interesting test-bed for interactive Natural Language Processing (NLP) systems and are also directly useful in a wide range of applications ranging from technical support services to entertainment. However, building intelligent conversational agents remains an unsolved problem in artificial intelligence research. Recently, recurrent neural network based models with sufficient capacity and access to large datasets attracted a large interest when first attempted. \citet{vinyals_neural_2015} showed that they were capable of generating meaningful responses in some chit-chat settings. Still, further inquiries in the capabilities of these neural network architectures and developments \citep{serban_hierarchical_2016,miao_neural_2015,sordoni_hierarchical_2015,serban_deep_2017,li_simple_2016,li_adversarial_2017} indicated that they were limited which made communicating with them a rather unsatisfying experience for human beings.

The main issues with these architectures can be summarized as:
\begin{itemize}
\item (i) the wildly inconsistent outputs and the lack of a consistent personality \citep{li_neural_2016},
\item (ii) the absence of a long-term memory as these models have difficulties to take into account more than the last dialogue utterance; and 
\item (iii) a tendency to produce consensual and generic responses (e.g. “I don’t know”) which are vague and not engaging for humans \citep{li_simple_2016}.
\end{itemize}

In this work, we make a step toward more consistent and relevant data-driven conversational agents by proposing a model architecture, associated training and generation algorithms which are able to significantly improve over the traditional seq-2-seq and information-retrieval baselines in terms of (i) relevance of the answer (ii) coherence with a predefined personality and dialog history, and (iii) grammaticality and fluency as evaluated by automatic metrics.

\section{Tasks and evaluation}

An interesting challenge to evaluate the quality of open-domain conversation agent is the Conversational Intelligence Challenge 2\footnote{\url{http://convai.io/}} (ConvAI2) that was held during the NIPS 2018 conference and which we shortly present here with its associated dataset.

ConvAI2 is based on the {\sc persona-chat} dataset \citep{zhang_personalizing_2018}, a crowd-sourced dialogue dataset in which each speaker was asked to condition its utterances on a predefined profile comprising a few sentences defining a personality as illustrated on figure \ref{table:persona-chat-example}. Paired workers were asked to chat naturally and to get to know each other during the conversation. This produced an interesting dataset with rapid turns of topics as it can be seen on the example we reproduce on table \ref{table:persona-chat-example}.

\begin{table*}[t]
  \begin{center}
    \begin{small}
      \begin{tabular}{l|l}
        \hline
        \textbf{Persona 1} & \textbf{Persona 2}\\
        \hline
I like to ski & I am an artist\\
My wife does not like me anymore & I have four children\\
I have went to Mexico 4 times this year & I recently got a cat \\
I hate Mexican food &  I enjoy walking for exercise \\
I like to eat cheetos &  I love watching Game of Thrones\\
\hline
\multicolumn{2}{l}{ }\\
\multicolumn{2}{l}{[PERSON 1:] Hi}\\
\multicolumn{2}{l}{[PERSON 2:] Hello ! How are you today ?}\\
\multicolumn{2}{l}{[PERSON 1:] I am good thank you , how are you.}\\
\multicolumn{2}{l}{[PERSON 2:] Great, thanks ! My children and I were just about to watch Game of Thrones. }\\
\multicolumn{2}{l}{[PERSON 1:] Nice ! How old are your children?}\\
\multicolumn{2}{l}{[PERSON 2:] I have four that range in age from 10 to 21. You?}\\
\multicolumn{2}{l}{[PERSON 1:] I do not have children at the moment.}\\ 
\multicolumn{2}{l}{[PERSON 2:] That just means you get to keep all the popcorn for yourself.}\\
\multicolumn{2}{l}{[PERSON 1:] And Cheetos at the moment!}\\
\multicolumn{2}{l}{[PERSON 2:] Good choice. Do you watch Game of Thrones?}\\
\multicolumn{2}{l}{[PERSON 1:] No, I do not have much time for TV.}\\
\multicolumn{2}{l}{[PERSON 2:] I usually spend my time painting: but, I love the show.}\\
      \end{tabular}
      \caption{Example dialog from the {\sc persona-chat} dataset. Person 1 is given their own persona (top left) at the beginning of the chat, but does not know the persona of Person 2, and vice-versa. They have to get to know each other during the conversation.
 \label{table:persona-chat-example}}
    \end{small}
  \end{center}
\end{table*}

As automatic evaluation is still an open question in dialogue systems \citep{liu_how_2016}, the {\sc persona-chat} dataset comes with three automated metrics on its evaluation set. The ConvAI2 challenge further evaluated these metrics on a privately held portion of {\sc persona-chat} combined with human evaluation.

The automatic metrics involves three tasks defined on the same dataset which are (i) \textit{a language modeling task} where the metric is the perplexity of gold utterance tokens as computed from the model's next token probability predictions (denoted \textbf{PPL}) (ii) a \textit{next utterance retrieval task} where the associated metric is the accuracy of retrieving a gold next utterance among 19 random distractor responses sampled from other dialogues (denoted \textbf{Hits@1}) and (iii) \textit{a generation task} which consists in generating a response in the dialog setting and where the metric is the F1 (precision and recall) of the content words of a gold dialog utterance in the predicted utterances (denoted \textbf{F1}).

Human evaluations are based on a combination of four metrics: fluency, consistency, engagingness (each evaluated as a grade between 1 and 5) and whether the human could guess the persona used by the bot (selection between two possible personas).

\section{Model}
The generative model used in \textbf{TransferTransfo} is a multi-layer Transformer encoder based on the Generative Pre-trained Transformer of \citeauthor{radford_improving_2018}. This model largely follows the original transformer work of \citeauthor{vaswani_attention_2017}. For more details on this recent model architecture which has become ubiquitous in Natural Language Processing, we refer readers to the detailed guide recently published by the Harvard SEAS natural-language processing group: “The Annotated Transformer.”\footnote{\url{http://nlp.seas.harvard.edu/2018/04/03/attention.html}}

We used a 12-layer decoder-only transformer with masked self-attention heads (768 dimensional states and 12 attention heads). By masked attention, we mean that the Transformer uses constrained self-attention where every token can only attend to its left context. In the literature this version of the Transformer is often referred to as a “Transformer decoder” since it is similar to the decoder part of the original encoder-decoder Transformer of \citet{vaswani_attention_2017}.

This model is similar to the large Transformer model recently used in several works leading to impressive results on several down-stream NLP tasks \citep{radford_improving_2018,devlin_bert:_2018}. Our model is based on a recently published PyTorch adaptation by the HuggingFace team which can be found at: \url{https://github.com/huggingface/pytorch-openai-transformer-lm}.

Following \citeauthor{radford_improving_2018,devlin_bert:_2018} the model uses learned positional embeddings with supported sequence lengths up to 512 tokens. The input sentences are pre-processed and tokenized using bytepair encoding (BPE) vocabulary with 40,000 merges \citep{sennrich_neural_2015}.

\section{Training}
\subsection{Pre-training}
Following the work of \citeauthor{radford_improving_2018}, the model is pre-trained on the BooksCorpus dataset \citep{zhu_aligning_2015} which contains over 7,000 unpublished books (about 800M words) from a variety of genres (Adventure, Fantasy, Romance...). The critical choice for this pre-training dataset is to use a document-level corpus rather than a shuffled sentence-level corpus to take advantage of long contiguous sequences and paragraphs and learn to condition on long-range information. This is not possible with shuffled sentence-level corpora such as the Billion Word Benchmark \citep{chelba_one_2013} used for instance in ELMo \citep{peters_deep_2018}. We used the pre-trained model weights open-sourced by \citeauthor{radford_improving_2018}.

\subsection{Fine-tuning}
After the pre-training step, the model is fine-tuned on the {\sc persona-chat} dataset using an augmented input representation and a multi-task learning scheme that we will now describe in greater details.
\begin{figure*}
	\includegraphics[width=\textwidth]{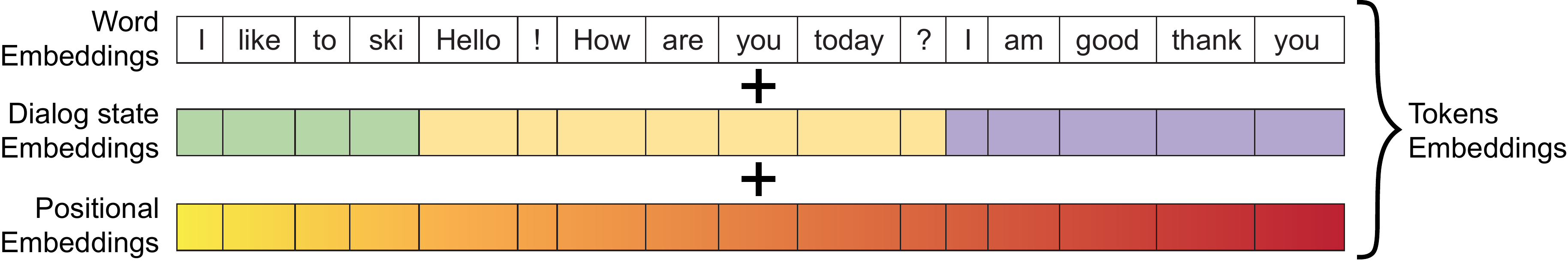}
    \caption{TranferTransfo's input representation. Each token embedding is the sum of a word embedding, a dialog state
embedding and a positional embedding.}
    \label{fig:inputs}
\end{figure*}

\subsubsection{Input representation}
We adapt the input representation of the model to be able to switch from a single (or unannotated) speaker setting like the one of the BookCorpus dataset to a two-speakers settings plus personality sentences like the one of the {\sc persona-chat} dataset.

More precisely, a sequence of input tokens for the model is constructed for each utterance by concatenating all the persona sentences of the current speaker (usually 4 to 6 sentences in the {\sc persona-chat} dataset) with a history of the dialog's previous utterances (typically 3 to 5 previous utterances).

From this sequence of input tokens, a sequence of input embeddings for the Transformer is constructed as follows. The word and positional embeddings learned during the pre-training phase are augmented with a set of dialog-state embeddings illustrated on figure \ref{fig:inputs}.

This set of additional embeddings is used to indicate whether the current token is part of (i) a personality sentence, (ii) an utterance from PERSON1 or (iii) an utterance from PERSON2.
These additional embeddings are learned on the {\sc persona-chat} dataset during the fine-tuning phase.

The input of the self-attention block of the Transformer model is then the sum of the three types of embeddings (word, dialog-state and positional) for each word.

Separation tokens may also be added to further separate each utterances of the dialog as it is commonly done for Transformer's inputs \citep{radford_improving_2018,devlin_bert:_2018}.

Another simple adaptation from pre-training to fine-tuning is to promote an invariance to personality sentence ordering by reusing the same positional embeddings for each personality sentences. This is similar in spirit to the Set Transformer recently proposed in \citeauthor{lee_set_2018}. Self-attention model are inherently insensitive to position and ordering and this feature can be conveniently harnessed to bias toward positional invariance. One interesting invariance that can be observed in conditional dialog datasets like the {\sc persona-chat} dataset is the invariance of the predicted utterances with respect to various orders of the personality sentences conditioning the dialog. A similar effect can be obtained by augmenting the training dataset with copies of the dialogs wherein the personality sentences are shuffled.

\subsubsection{Multi-task learning}
Fine-tuning is done by optimizing a combination of two loss functions: (i) a next-utterance classification loss, and (ii) a language modeling loss.

\begin{figure*}
	\includegraphics[width=\textwidth]{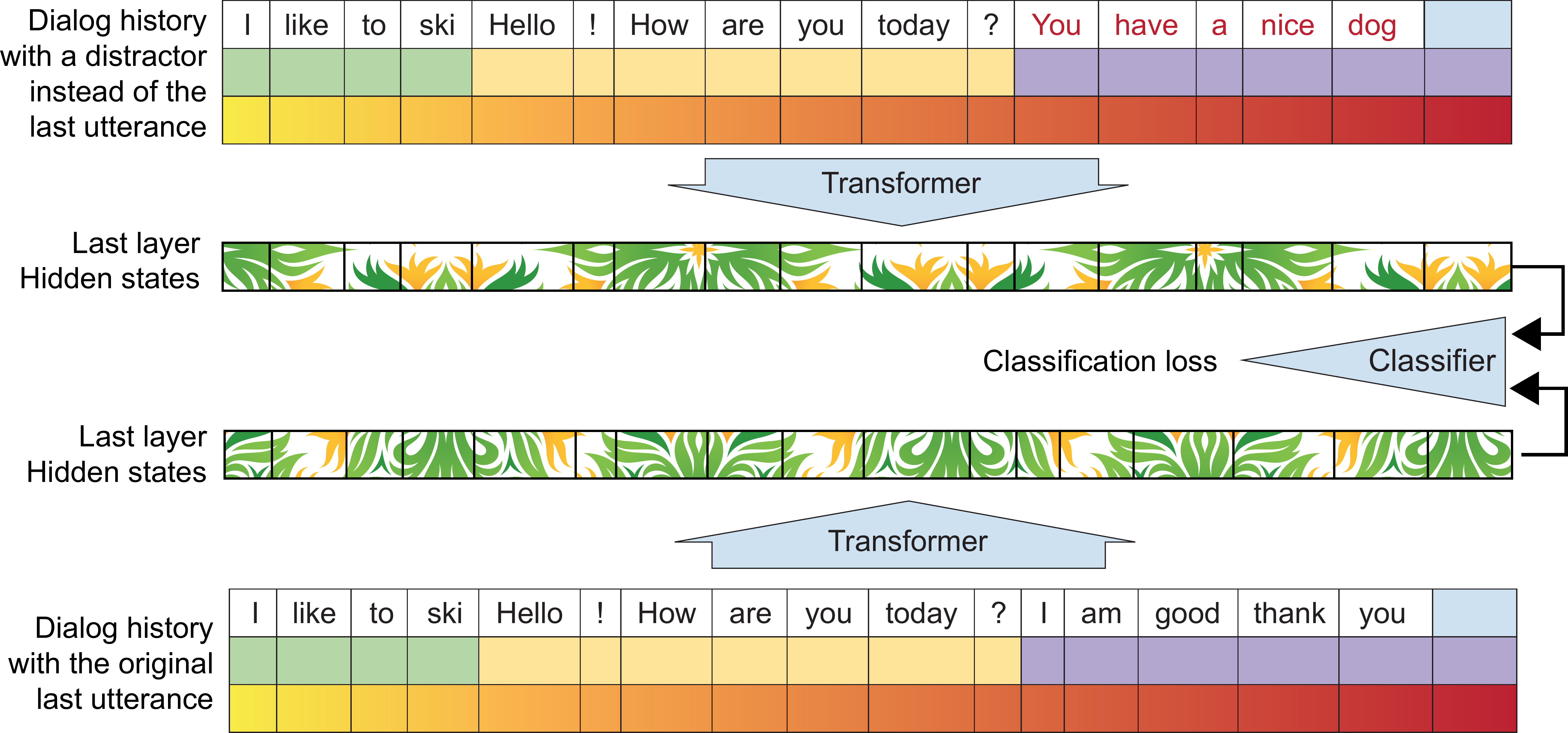}
    \caption{TranferTransfor input representation. The input embeddings is the sum of the word embeddings, the dialog state
embeddings and the positional embeddings.}
    \label{fig:classif}
\end{figure*}

The \textit{next-utterance classification loss} is illustrated on figure \ref{fig:classif} and bears similarities with the Next Sentence Prediction task developed in a parallel work by \citeauthor{devlin_bert:_2018}. It consists in training a classifier to distinguish a correct next utterance appended to the input sequence from a set of randomly sampled distractors (in practice between 2 and 6 randomly sampled utterances). The classifier is a linear layer taking as input the last hidden state of the self-attention model and computing a score. For classification a special token $[CLS]$ is added at the sentence illustrated in blue on figure \ref{fig:classif}, the last hidden state used for the classifier thus corresponds to the hidden-state associated to this termination special token. The computed scores are passed through a softmax layer to obtain classification probabilities. The parameters of the Transformer and the next-utterance classifier layer are fine-tuned jointly to maximize the log-probability of the correct label.

The \textit{language modeling loss} is the commonly used cross-entropy loss where the final hidden state of the self-attention model is fed into an output softmax over the vocabulary to obtain next token probabilities. These probabilities are then scored using a negative log-likelihood loss where the gold next tokens are taken as labels.

\begin{table*} [b]
\centering
\resizebox{2\columnwidth}{!}{%
\begin{tabular}{|l || c|c|c || c|c|c |}
\hline
& \multicolumn{3}{c||}{Eval} & \multicolumn{3}{c|}{Test} \\
\hline
\textbf{Model} & PPL & Hits@1 & F1 & PPL & Hits@1 & F1  \\
 \hline
  Generative Profile Memory \citep{zhang_personalizing_2018} & \textit{34.54} & 12.5 & -- & -- & -- & --	 \\ 
  Retrieval KV Profile Memory \citep{zhang_personalizing_2018} & -- & \textit{51.1} & & & &	 \\ 
  Seq2Seq + Attention (ConvAI2 baseline\footnote{\url{http://convai.io/}})  & 35.07 & 12.5 & \textit{16.82} & \textit{29.8} & 12.6 & \textit{16.18} \\ 
  Language Model (ConvAI2 baseline\footnote{\url{http://convai.io/}})  & 51.1 & -- & 15.31 & 46.0 & -- & 15.02 \\ 
  KV Profile Memory (ConvAI2 baseline\footnote{\url{http://convai.io/}})  & -- & 55.1 & 11.72 & -- & \textit{55.2} &	11.9 \\ 
\hline
  TransferTransfo (this work) & \textbf{17.51} & \textbf{82.1} & \textbf{19.09} & \textbf{16.28} & \textbf{80.7} & \textbf{19.5}\\ 
\hline
\end{tabular}%
}
\caption{Results on the (public) validation and (private) test set of the {\sc persona-chat} dataset. The results on the test set were evaluated by the ConvAI evaluation server. PPL stands for perplexity, Hits@1 for correct identification of a gold answer from a set of 19 distractors and F1 for precision and recall of content words in a dialog utterance (see \citeauthor{zhang_personalizing_2018} and \url{http://convai.io/} for details)}
\label{tab:result-comparisons}
\end{table*}

\subsubsection{Fine-tuning details}

We fine-tuned the model with a batch size of 32 sequences having an average of 250 tokens depending on the batch for 200,000 steps, which is approximately 2 epochs over the {\sc persona-chat} training dataset (32 sequences * 250 tokens = 8,000 tokens/batch). We used Adam with a learning rate of 6.25e-5, $\beta_1 = 0.9$, $\beta_2 = 0.999$, L2 weight decay of 0,01 and a coefficient of 2 on the Language Modeling loss when summing with the next-utterance classification loss losses. The learning rate was linearly decayed to zero over the course of the training. We use a dropout probability of 0.1 on all layers. Following \citeauthor{radford_improving_2018} we use a $relu$ activation function. Fine-tuning the model took about 10h on four K80 GPUs.

\subsubsection{Decoding details}

Generation was performed using beam search with sampling and a small beam size of 4. Simple n-grams filtering is used to ensure the model doesn't directly copy from the personality sentences (forbidden by the ConvAI2 rules) as well as older utterances. The final beams are ranked according to a scalar combination of the length-normalized utterance probability and the next-utterance classification score. Increasing the importance of the next-utterance classification score results in utterances that stick more closely to the provided personality sentences but also reduce the diversity of the dialog.

\section{Results}

Results on the public evaluation split and the privately-held test splits of the {\sc persona-chat} dataset are illustrated on table \ref{tab:result-comparisons}. TransferTransfo outperforms the existing systems by a significant margin on the public validation dateset obtaining $51\%$ absolute improvement in perplexity (PPL), $35\%$ absolute improvement in Hits@1 and $13\%$ improvement in F1.

More importantly, while the model's hyper-parameters were tuned on the validation set, the performance improvements translate to the private test set as scored by the ConvAI2 evaluation server with a $45\%$ absolute improvement in perplexity (PPL), $46\%$ absolute improvement in Hits@1 and $20\%$ improvement in F1.

The perplexity is noticeably low for an open-domain language modeling task which may be in-part due to a few repetitive portions of the dataset like the introductory utterances at the beginning of each dialog ("Hello, how are you?") and the copy mechanisms from the personality sentences.

\section{Conclusion}

Transfer learning from language models have been recently shown to bring strong empirical improvements in discriminative language understanding tasks. In the present work, we show that such improvements can be extended to generative tasks such as open-domain dialog generation which combine many linguistics aspects such as co-reference resolution, common-sense knowledge and long-range dependency modeling among others. We offer hints as to what kind of multi-task fine-tuning setups can be effective in these setups and illustrate the effectiveness of this approach on a recent dialog task. Important future work is still needed to understand the most optimal settings and models.

\bibliography{Zotero}
\bibliographystyle{aaai}

\end{document}